\documentclass[runningheads]{llncs}

\usepackage[T1]{fontenc}

\usepackage{amsfonts}
\usepackage{amsmath}
\usepackage{graphicx,verbatim}

\usepackage{hyperref}
\usepackage{color}

\usepackage{booktabs}
\usepackage{nicefrac}

\begin{document}

\title{Leveraging generic foundation models for multimodal surgical data analysis}

\author{Simon Pezold\inst{1,2}\orcidID{0000-0002-4998-8983} \and
Jérôme A. Kurylec\inst{1,2}\orcidID{0009-0000-8999-9110} \and
Jan S. Liechti\inst{1,2} \and
Beat P. Müller\inst{1,2}\orcidID{0000-0002-8552-8538} \and\\
Joël L. Lavanchy\inst{1,2}\orcidID{0000-0003-0248-4996}}

\authorrunning{S. Pezold et al.}

\institute{Department of Biomedical Engineering, University of Basel, Allschwil, Switzerland \and
Clarunis – University Digestive Health Care Center Basel, Basel, Switzerland
\email{simon.pezold@unibas.ch}}

\maketitle

\begin{abstract}
We investigate how both the adaptation of a generic foundation model via transfer learning and the integration of complementary modalities from the operating room can support surgical data science. To this end, we use V-JEPA, which has been trained on natural video scenes, as the single-modality foundation of a multimodal model for minimally invasive surgery support. We analyze how the model’s downstream performance can benefit (a)~from finetuning on unlabeled surgical video data and (b)~from providing additional time-resolved data streams from the operating room in a multimodal setup.

In an in-house dataset of liver surgery videos, we analyze the tasks of predicting hospital length of stay and postoperative complications. In videos of the public HeiCo dataset, we analyze the task of surgical phase recognition. As a baseline, we apply pretrained V-JEPA to all tasks. We then finetune V-JEPA on unlabeled, held-out videos to investigate the model’s change in performance after domain adaptation. Following the idea of modular decision support networks, we integrate additional data streams from the operating room by training a separate encoder to form a shared representation space with V-JEPA’s embeddings.

Our experiments show that finetuning on domain-specific data increases model performance. On the in-house data, we find that integrating additional time-resolved data likewise benefits the model. On the HeiCo data, accuracy of the pretrained video-only, single-modality baseline setup is on par with the top-performing submissions of the EndoVis2017 challenge, while finetuning on domain-specific data increases accuracy further. Our results thus demonstrate how surgical data science can leverage publicly available generic foundation models. Likewise, they indicate the potential of domain adaptation and of integrating suitable complementary data streams from the operating room for improved model performance. To support further research, we release our code and model weights at \url{https://github.com/DigitalSurgeryLab-Basel/ML-CDS-2025}.


\end{abstract}

\section{Introduction}

In recent years, the release of foundation models on various modalities such as text, image, or video data has enabled a growing range of machine learning (ML)–based applications. Foundation models can serve as building blocks for a wide range of downstream tasks, enabling their users to leverage the benefits of large-scale pretraining, which is potentially faster and less resource-intensive than designing and training complete models from scratch. In very specialized areas, such as medical applications, however, the question may arise as to whether foundation models that have been trained on generic real-world data are applicable to the target domain.

Multimodality, that is, the simultaneous use of various modalities as data sources in a single model, is likewise an increasingly popular approach in the creation of ML models. The main potential of such a setup is its ability to reinforce relevant signals and close informational gaps from one specific modality alone. Recent work has demonstrated the potential of multimodal models, especially vision--language models (VLMs), to advance surgical applications. SurgVLP \cite{yuan_learning_2025}, for example, leverages publicly available sources of surgical video lectures to learn joint vision--language representations suitable for surgery-specific downstream tasks, using a self-supervised learning objective. SSG-VQA-Net \cite{yuan_advancing_2024} employs a scene graph--based dataset for visual question answering (VQA) on surgical scenarios. GP-VLS \cite{schmidgall_gp-vls_2024} extracts knowledge from medical textbooks, question--answer pairs, and text-enriched surgical scenes for general-purpose VQA in surgery. Khan et al. \cite{khan_comprehensive_2025} provide a recent survey of foundation models in medicine, showing that such models, if multimodal, mostly combine images and text or purely data from different imaging modalities. Thus, while multimodal setups have already been established in medical research, their use with data from the operating room (OR), especially regarding the combination of visual information with sources other than language, has been underexplored.

\begin{figure}[t!]
\includegraphics[width=\textwidth]{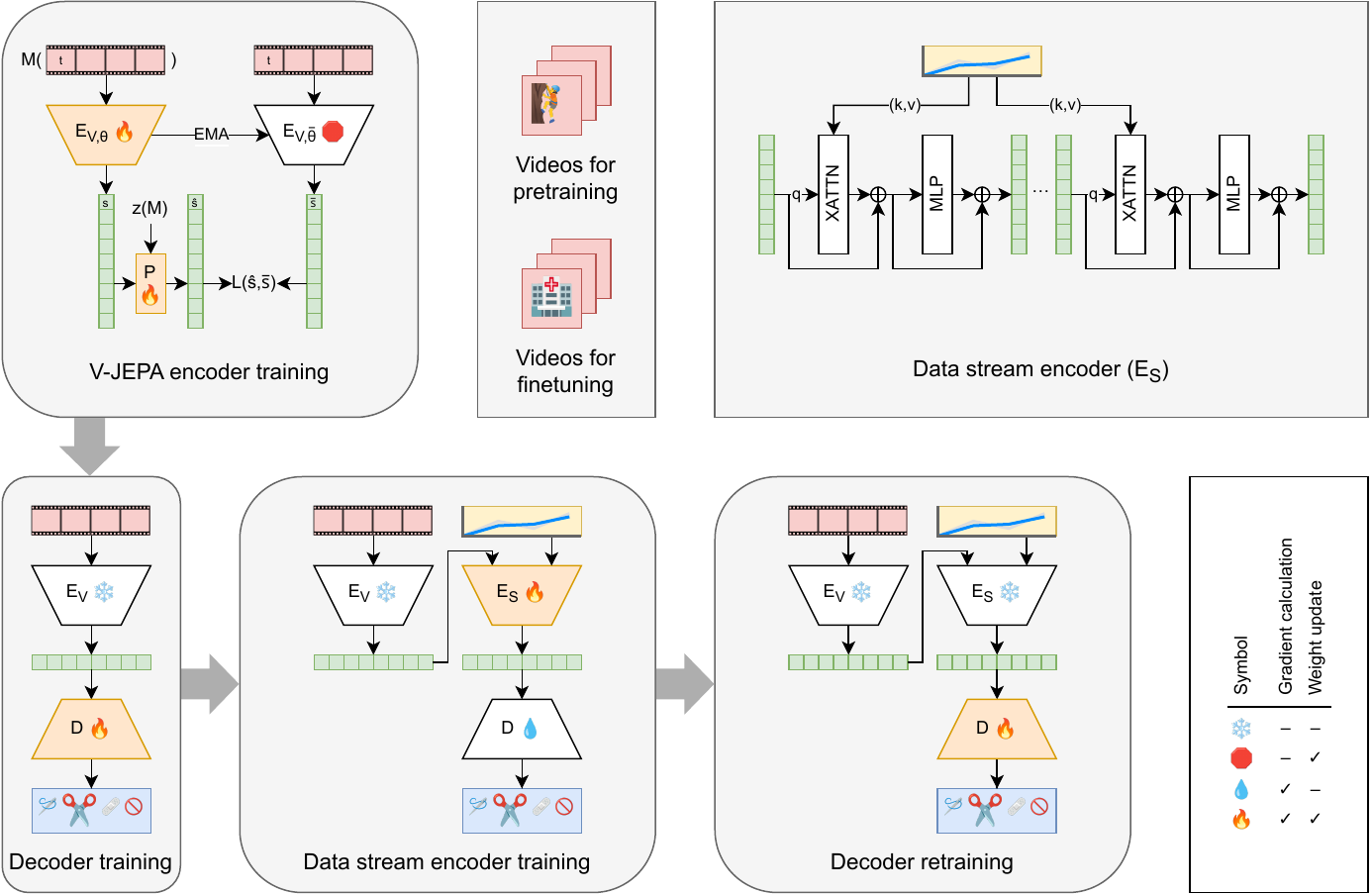}
\caption{Architecture and training steps. \emph{V-JEPA training step (top left):} Video encoder $E_{V,\theta}$ is trained by producing an embedding $s$ \emph{(green)} of the unmasked tokens from partially masked, tokenized video data $M(t)$ \emph{(red)}, which enables auxiliary predictor $P$ to produce an estimate $\hat{s}$ of the masked video tokens' embedding $\bar{s}$, provided by exponential moving average (\textsc{ema}) teacher $E_{V,\bar{\theta}}$ (see main text for details). \emph{Used data (top center):} Pretraining $E_V$ relies on clips from everyday scenes, finetuning $E_V$ employs endoscopic clips from the OR. \emph{Data stream encoder architecture (top right)}: Over a sequence of layers, stream data \emph{(yellow)} and embeddings are fused via cross-attention (\textsc{xattn}) to update the latter, where the queries $q$ are formulated on previous embeddings and the keys and values $(k,v)$ are formulated on the stream data, followed by a multi-layer perceptron (\textsc{mlp}). \emph{Decoder training step (bottom left):} Decoder $D$ is trained on its downstream task \emph{(blue)}, receiving embeddings of a frozen $E_V$. \emph{Data stream encoder training step (bottom center):} Stream encoder $E_S$ is trained via $D$'s downstream task by receiving embeddings from a frozen $E_V$ and gradients from the trained $D$, the weights of which are not updated. \emph{Decoder retraining step (bottom right):} Training of $D$ is continued on its downstream task, now receiving embeddings from the frozen $E_S$.} \label{fig-overview}
\end{figure}

\subsubsection{Contribution.} The aim of this work is twofold: First, we analyze whether a publicly available foundation model for processing generic real-world video data is suitable for use in surgical video analysis. Second, we investigate whether medical downstream tasks can benefit from the integration of additional data sources from the OR. To this end, we hypothesize (1)~that a generic foundation model can form the basis for a domain-specific downstream task, (2)~that domain adaptation can increase model performance, and (3)~that the inclusion of additional data sources can increase model performance even further.

In a series of experiments laid out in Section~\ref{experiments}, we challenge our hypotheses on the basis of an in-house and a public dataset: We use pretrained V-JEPA~v1 from Bardes et al.~\cite{bardes_revisiting_2024} as a baseline encoder to train two task heads from its embeddings on the in-house dataset, one for predicting prolonged hospital length of stay, one for predicting increased complication severity. Using the same baseline encoder and training another task head on the publicly available HeiCo dataset, our results in surgical phase recognition accuracy are on par with the top-performing competitors of the MICCAI 2017 endoscopic vision (EndoVis2017) challenge\footnote{We report challenge details and competitor results as presented at\\\url{https://endovissub2017-workflow.grand-challenge.org/PastChallenges/}.}. Next, we finetune V-JEPA in a self-supervised setting on held-out, unlabeled surgical videos and repeat the previous experiments, gaining overall improved predictions on the in-house data and surpassing reported accuracy results on HeiCo. Finally, we integrate additional time-resolved data streams (vital data for the in-house dataset, surgical device status for HeiCo) via an additional encoder, which likewise tends to increase model performance compared to baseline. To support further research in the chosen direction, we release our models and model weights at \url{https://github.com/DigitalSurgeryLab-Basel/ML-CDS-2025}.

\section{Methods and Materials} \label{method}

\subsubsection{V-JEPA.} As the basis for our model, we use pretrained V-JEPA~v1 \cite{bardes_revisiting_2024} (see Fig.~\ref{fig-overview}, top left, for a simplified schematic), the details of which we briefly reiterate here for the reader's convenience (throughout the paper, we use the name \emph{V-JEPA} to refer to either the model itself or its training, depending on the context).

At its core, V-JEPA uses a vision transformer (ViT) \cite{dosovitskiy_image_2021} with an extension to the temporal dimension. V-JEPA's published weights have been trained on 2~million videos of primarily human everyday activities without special focus on the medical field. Training has been achieved in self-supervised fashion on these videos, following the joint embedding predictive architecture (JEPA) paradigm \cite{yann_lecun_path_2022} and ignoring any potential manual labels. More specifically, in the V-JEPA training task, a student model $E_{V,\theta}$ produces embeddings $s$ from partially masked, tokenized input $M(t)$ \cite{he_masked_2022}, while a teacher model $E_{V,\bar{\theta}}$ receives the full video token sequence $t$ to produce its embedding $\bar{s}$. An auxiliary predictor $P$, which is discarded after training, predicts $\bar{s}$ by estimating $\hat{s}$ from $s$ for the masked tokens, receiving information about the masked tokens' locations as additional input $z(M)$. Gradients from the loss term $L(\hat{s},\bar{s})$, the latter being formulated as an $L_1$ loss, are used to update the student's weights $\theta$ as well as $P$'s weights through backpropagation, while the teacher's weights $\bar{\theta}$ are updated as an exponential moving average over the iterations on $\theta$. As the final training result, either $E_{V,\theta}$ or $E_{V,\bar{\theta}}$ can be chosen. We opt for the latter in our setup, following the experiments of Bardes et al.~\cite{bardes_revisiting_2024}.

\subsubsection{Multimodal extension.} In our aim to integrate additional modalities, we take inspiration from modular decision support networks \cite{swamy_multimodnmultimodal_2023,trottet_modular_2023}. In this setup, a randomly initialized state vector is successively consumed and updated by modality-specific encoders, enabling them to enrich the state with complementary information, which can later be decoded by task-specific heads \cite[Fig.~1]{swamy_multimodnmultimodal_2023}. The architectures of encoders and decoders, as well as the order of encoders, can be chosen freely in principle, thus allowing for great flexibility in general and for missing inputs in particular; namely, by skipping the corresponding encoder in the latter case if necessary.

Our adaptation (Fig.~\ref{fig-overview}, bottom center) slightly deviates from this approach, by using V-JEPA's embeddings as the initial state vector, implicitly assuming them as given. While therefore losing a part of the previously described flexibility, we believe this is outweighed by the benefit of being able to plug in the pretrained V-JEPA encoder as-is: this direct integration would otherwise not be possible, since V-JEPA has not been designed to receive a state vector as an additional input. Our adaptation, in this sense, also agrees with the locked-image tuning (LiT) approach of Zhai et al.~\cite{zhai_lit_2022}, where the authors find that freezing an image data encoder and aligning an additional encoder for multimodal integration (text data in their case, temporal data streams in ours) achieves favorable results.

For the actual integration of further modalities, we follow the idea of HEALNet \cite{hemker_healnet_2024} to design the corresponding encoders (Fig.~\ref{fig-overview}, top right). In particular, cross-attention is used to formulate the queries on the state vector produced by the previous modality, while the keys and values are formulated on the tokens of the current modality, thereby successively updating the state vector over the given modalities and predefined fusion layers. Just like HEALNet, for a smaller model footprint, we use shared weights for the linear transforms producing the keys, queries, and values, respectively, throughout all layers of the corresponding encoders.

\subsubsection{Architecture.} For the initial video encoder, we adopt the V-JEPA architecture as given. Likewise, for the task-specific decoder heads, we adopt the architecture of the attentive classifiers used in the V-JEPA experiments of Bardes et al.~\cite{bardes_revisiting_2024}, only adjusting the numbers of classes as necessary. For the HEALNet-based encoders, we proceed as follows:

Let $\vartheta_m: x_{m} \mapsto t_{m} \in \mathbb{R}^{n_m \times d}$ be a tokenizer that produces tokenized $(n_m \times d)$-dimensional samples $t$ from the raw samples $x$ (sample indices are omitted for brevity) of the $m$-th modality ($m=1,...,M$). Let $\varphi_m: (t_m,s_{m-1,\ell}) \mapsto s_{m,\ell} \in \mathbb{R}^{n_\mathrm{s} \times d}$ be the part of the $\ell$-th fusion layer ($\ell \in 1,...,L$) that updates the $(n_\mathrm{s} \times d)$-dimensional state $s$ with the tokenized samples $t$ for the $m$-th modality.

As already mentioned, we assume the outputs of V-JEPA as the initially given state $s_{1,1}$. For each successive modality ($m > 1$), we define $\varphi$ as
\begin{gather}
    \varphi_m(t_m,s_{m-1,\ell}) = \tilde{s} + \psi\left(\left|\tilde{s}\right|\right)\mbox{\quad with}\label{eq-fusion}\\
    \tilde{s} = s_{m-1,\ell} + W^\mathrm{o}_m\,\alpha\left( q, k, v \right) + b^\mathrm{o}_m\mbox{\quad and}\\
    q = W^\mathrm{q}_m\,s_{m-1,\ell} + b^\mathrm{q}_m,\quad k = W^\mathrm{k}_m\,\left| t_m \right| + b^\mathrm{k}_m,\quad v = W^\mathrm{v}_m\,\left| t_m \right| + b^\mathrm{v}_m\label{eq-qkv},
\end{gather}
where $\left|\cdot\right|$ signifies layer norm, $\psi(\cdot)$ is a two-layer perceptron with an intermediate GELU activation, and $\alpha(q, k, v)$ is scaled dot product attention over the given queries $q$, keys $k$, and values $v$. The final state of each fusion layer forms the initial state of the next layer ($s_{1,\ell}=s_{M,\ell-1}$).

With the chosen encoder components, we again take inspiration from the attentive classifiers in the V-JEPA experiments of Bardes et al.~\cite{bardes_revisiting_2024}, albeit with arbitrary depth and HEALNet-based cross-attention and weight sharing $\left(W^{\boldsymbol{\cdot}}_m, b^{\boldsymbol{\cdot}}_m\right)$, as already mentioned. Bardes et al.'s attentive classifiers, in turn, largely follow the original transformers design \cite{vaswani_attention_2017}.

\subsubsection{Training.} To achieve a stable training of the described components, we propose the following four-step training recipe:

\begin{enumerate}
\item \emph{V-JEPA encoder pretraining and finetuning (Fig.~\ref{fig-overview}, top left):} Use the V-JEPA training scheme with the self-supervised task of predicting embeddings, as outlined by Bardes et al. \cite{bardes_revisiting_2024} and as summarized above. For pretraining, rely on the published V-JEPA weights\footnote{Links to the weights can be found at \url{https://github.com/facebookresearch/jepa}.}. For finetuning, continue training on unlabeled videos from the domain of interest (endoscopic videos from surgery, in our case). In all following steps, keep the V-JEPA encoder frozen.

\item \emph{Decoder training (Fig.~\ref{fig-overview}, bottom left):} Train the decoder on its downstream task, using the initial state vector (that is, the V-JEPA embedding) as its input and employing a suitable loss term that integrates corresponding task labels from the labeled training data.

\item \emph{Encoder training for additional modalities (Fig.~\ref{fig-overview}, bottom center):} Continue calculating gradients on the decoder, but do not update its weights. Train the new modality's encoder on the decoder's downstream task, keeping the loss term from step~2, but combining it with a loss term that penalizes changes to the values in the state vector (or, in other words, that minimizes the difference between the V-JEPA embeddings and the new encoder's output).

\item \emph{Decoder retraining (Fig,~\ref{fig-overview}, bottom right):} Freeze the new encoder's weights. Continue training the decoder on its downstream task, now using the updated state vector (that is, the new encoder's output) as its input.
\end{enumerate}

\noindent While the choices of steps~1 and steps~2 arguably appear straightforward, steps~3 and 4 might require further motivation:

It is the task of step~3 to produce an updated state vector that incorporates the information of the new modality. As we cannot expect having labels for the new encoder's output and thus cannot easily formulate a loss term, we use the gradients backpropagated through the decoder as a surrogate signal for the encoder's weight updates.\footnote{Since the decoder, in this step, assumes what can be seen as a \emph{fluid} state (neither \emph{hot}, because its weights are not receiving updates, nor \emph{frozen}, as it stays part of the gradient flow), we use the water droplet symbol in Figure~\ref{fig-overview}.} As we furthermore want to align the new embeddings with those of the V-JEPA encoder, we rely on the state change penalty \cite{trottet_modular_2023}. Other formulations could be used here; for example, a loss term solely based on the state change penalty or predicting the V-JEPA embedding from the new encoder's embedding via an auxiliary predictor, similar to the V-JEPA training task. However, since the V-JEPA embedding is an input to the encoder, we did not explore whether meaningful outputs could indeed be produced this way.

Step~4 can be seen as a consequence from step~3: since the state vector should now contain information from the additional modality, we try to ensure that this information can be put to the best use by the decoder, rather than interfering with it, by allowing the decoder to adjust for the new input values.

\subsubsection{Datasets.} The following paragraphs summarize the content of the used datasets and their role in our experiments. A graphical overview of the datasets is provided in the \nameref{sec-app}.

\paragraph{HeiCo dataset:} The Heidelberg colorectal (HeiCo) dataset \cite{maier-hein_heidelberg_2021} provides laparoscopic videos from three types of surgery: proctocolectomy (10~videos, $34{:}02\,\mbox{h}$), rectal resection (10~videos, $35{:}22\,\mbox{h}$), sigmoid resection (10~videos, $27{:}04\,\mbox{h}$). As an additional modality, for each video frame, 14~status values from medical devices in the OR are provided: insufflator (current flow, target flow, current pressure, target pressure, used volume, supply pressure, on/off), OR lights (on/off, intensity 1, intensity 2), endoscope light (intensity), endoscope camera (white balance, gains, exposure). As an outcome label, for each video frame, one of 14~surgical phases has been annotated: \emph{general preparation and orientation in the abdomen~(0), dissection of lymph nodes and blood vessels en bloc~(1), retroperitoneal preparation towards lower pancreatic border~(2), retroperitoneal preparation of duodenum and pancreatic head~(3), mobilization of sigmoid colon and descending colon~(4), mobilization of splenic flexure~(5), mobilization of transverse colon~(6), mobilization of ascending colon~(7), dissection and resection of the rectum~(8), extra-abdominal preparation of anastomosis~(9), intra-abdominal preparation of anastomosis~(10), creation of stoma~(11), finalization of operation~(12), exception~(13)}.

In our experiments below, we use the HeiCo data with the splits provided for EndoVis2017. Moreover, we use the videos from the challenge's training split (15~videos, $46{:}35\,\mbox{h}$) and validation split (9~videos, $28{:}21\,\mbox{h}$) as part of the data for domain-specific finetuning of V-JEPA (step~1 in our training recipe).

\paragraph{In-house dataset:} Our in-house dataset consists of laparoscopic videos (55~videos, $148{:}14\,\mbox{h}$) from different patients (23~female, 32~male; age: $62.4\,\pm\,15.5\,\mbox{y}$, $21\,\mbox{y}$ min., $84\,\mbox{y}$ max.) undergoing minimally invasive liver surgery. As an additional modality, 4 streams of vital data (heart rate; diastolic, mean, systolic arterial blood pressure) have been recorded throughout the surgery. As an outcome label, the patients' length of stay in terms of postoperative hospital days ($4\,\mbox{d}$ median, $2\,\mbox{d}$ min., $14\,\mbox{d}$ max.; 13~patients above median) and complication impact in terms of the  comprehensive complication index (CCI) \cite{slankamenac_comprehensive_2013} ($0.0$ median, $0.0$ min., $36.2$ max.; 9~patients above median) have been recorded.

We use our in-house data in two series of experiments below, one for length of stay and one for CCI prediction.

\paragraph{Additional finetuning data:} In addition to parts of the HeiCo data (see above), we use all videos from the Cholec80 \cite{twinanda_endonet_2017} dataset (80~laparoscopic videos of cholecystectomy surgery, $51{:}15\,\mbox{h}$) and MultiBypass140 \cite{lavanchy_challenges_2024} dataset (laparoscopic videos of gastric bypass surgery; 70~videos, $87{:}56\,\mbox{h}$, from Inselspital Bern; 70~videos, $129{:}09\,\mbox{h}$, from University Hospital Strasbourg) for domain-specific finetuning of V-JEPA (step~1 in our training recipe).

\section{Experiments and Results} \label{experiments}

\subsubsection{Preliminaries.} The following paragraphs summarize steps and settings that apply to all experiments described below.

\paragraph{Pretrained vs. finetuned V-JEPA:} For all experiments, we use the V-JEPA ViT-L model size. By \emph{pretrained V-JEPA}, we refer to the weights provided by Bardes et al. \cite{bardes_revisiting_2024}. By \emph{finetuned V-JEPA}, we refer to the weights that we gained by continued self-supervised JEPA training of pretrained V-JEPA on the finetuning data (see the \emph{Datasets} section above), following step~1 of our training recipe.

Specifically, we continued training on a system of 4 Nvidia GH200 chips with 96\,GB shared CPU--GPU memory per chip, adopting the training schedule from the V-JEPA code base, albeit adjusting the batch size to 112 and the number of batches sampled per epoch to 1000, and reducing both weight decay (by a factor of 10) and learning rate (by a factor of 100) for finetuning. Due to resource restrictions, training was limited to 12~hours wall-clock time, which amounted to 7 completed epochs of continued training on the finetuning data.

\paragraph{Data stream encoder design:} Data stream encoders used throughout experiments adopted the previously described HEALNet-inspired architecture, with a depth of 4~layers, 4~heads in cross-attention, and a dropout rate of $0.2$ during training. The used encoders consume all streams at once but tokenize each stream separately, using the patch width from the video encoder, but applying it in a sliding window fashion to get a denser signal. For a stable initial learning behavior, residual connections in all layers were initialized as identity functions by zero-initializing the corresponding model parameters.

\begin{table}[t!]
\caption{Training hyperparameters . The \emph{schedule stretch factor} refers to the following observation of Bardes et al. \cite{bardes_revisiting_2024}: \emph{``We found the last 25\% of the default scheduler period to update hyperparameters too aggressively, and simply truncating the schedulers improved performance.''} For further details and values, see main text.}\label{tab-params}
\centering\setlength{\tabcolsep}{6pt}
\begin{tabular}{@{}lrrr@{}}
\toprule
 & Step 2, HeiCo & Step 2, in-house & Steps 3+4, all \\
\midrule
Epochs & 25 & 25 & 10 \\
Samples per epoch & 50,000 & 1000 & 1000  \\
Learning rate (start, max., end) & $10^{-3}$, $10^{-3}$, 0 & $10^{-5}$, $10^{-3}$, 0 & $10^{-6}$, $10^{-4}$, 0 \\
Weight decay (start, end) & $10^{-2}$, $10^{-6}$ & $10^{-2}$, $10^{-2}$ & $10^{-4}$, $10^{-4}$ \\
Schedule stretch factor & 100\% & 125\% & 125\% \\
\bottomrule
\end{tabular}
\end{table}

\paragraph{HeiCo data handling:} We followed the EndoVis2017 evaluation as closely as possible, filling in gaps in its evaluation protocol with reasonable assumptions. Specifically: We skipped all out-of-body segments, which have been replaced by all-blue video frames in the HeiCo data, for both training and evaluation. For evaluation, we densely sampled clips at one-second intervals from all videos in the test split. Regarding the class denoting exceptions in the surgical phase annotations (class~13), we report aggregated evaluation results both including it (for completeness) and excluding it (as in the challenge).

\paragraph{In-house data handling:} We created a setup that enabled a similar training and evaluation as with the HeiCo data. Specifically: (1)~We resampled the vital data streams to match the timestamps of the video frames. (2)~We modeled outcome prediction as a binary classification problem for both targets: \emph{value at or below median (class 0) vs. value above median (class 1)}. (3)~We resampled each patient's binary class label as a frame-wise value and densely sampled clips at one-second intervals from all videos in the test split for evaluation. (4)~We provided separate splits, ensuring a balanced distribution of classes, for the outcomes of length of stay (training: 29~videos, $50{:}38\,\mbox{h}$ class~0, $25{:}29\,\mbox{h}$ class~1; validation: 13~videos, $27{:}16\,\mbox{h}$ class~0, $9{:}21\,\mbox{h}$ class~1; testing: 13~videos, $25{:}49\,\mbox{h}$ class~0, $9{:}41\,\mbox{h}$ class~1) and CCI (training: 29~videos, $65{:}37\,\mbox{h}$ class~0, $14{:}38\,\mbox{h}$ class~1; validation: 13~videos, $28{:}43\,\mbox{h}$ class~0, $5{:}52\,\mbox{h}$ class~1; testing: 13~videos, $27{:}48\,\mbox{h}$ class~0, $5{:}36\,\mbox{h}$ class~1), respectively.

\paragraph{Training hyperparameters:} Our hyperparameters largely follow the V-JEPA code base: In all experiments, we used the AdamW optimizer. Linear warmup for 1~epoch was combined with cosine decay for the learning rate. Weight decay followed a cosine schedule without warmup. Patches were sampled with equal probability from all classes and supplied to the model with a batch size of 4. Classification errors were penalized via cross-entropy loss, changes to the state vector in the multimodal experiments via $L_2$ loss. In the latter case, both losses were summed with a relative weight of $10^{-3}$ on the state change penalty. Further hyperparameters for steps~2 to 4 of our training recipe are summarized in Table~\ref{tab-params}.

\begin{table}[t!]
\caption{Results on the HeiCo dataset. The top section quotes the top-3 competitors from the EndoVis2017 challenge. The bottom section presents results from our own experiments on the dataset. \emph{Pretrained} and \emph{finetuned} refer to the use of pretrained and finetuned V-JEPA, respectively, as defined in the main text. Shown metrics reproduce the ones from the challenge; these are: average IoU (aIoU), median IoU (mIoU), $\nicefrac{1}{6}$ quantile IoU (qIoU), overall accuracy (Acc). Numbers in brackets show results including class~13 (see main text for details).}\label{tab-heico}
\centering\setlength{\tabcolsep}{6pt}
\begin{tabular}{@{}lrrrr@{}}
\toprule
Approach & \%aIoU & \%mIoU & \%qIoU & \%Acc \\
\midrule
EndoVis \#1 (video-only) & \textbf{40} & \textit{38} & \textbf{37} & 61 \\
EndoVis \#2 (video+streams) & \textit{38} & \textit{38} & \textit{36} & 60 \\
EndoVis \#3 (video-only) & 25 & 25 & 25 & 57 \\ 
\midrule
pretrained (video-only) & 37 (34) & 32 (32) & 25 (15) & \textit{62} (60) \\
finetuned (video-only) & \textit{38} (35) & 35 (34) & 26 (12) & \textbf{64} (62) \\
pretrained (video+streams) & 37 (34) & 37 (35) & 17 (14) & 59 (58) \\
finetuned (video+streams) & \textit{38} (35) & \textbf{41} (39) & 21 (10) & \textit{62} (60) \\
\bottomrule
\end{tabular}
\end{table}

\subsubsection{Results on HeiCo data.} Table~\ref{tab-heico} shows results for the task of surgical phase recognition on the HeiCo dataset.

Regarding the use of domain-specific finetuning for the underlying V-JEPA video encoder, one can see its benefit in all metrics, both in the single-modality setup (corresponding to step~2 in our training recipe) and in the multimodal setup (corresponding to step~4). The benefit of integrating surgical device status as an additional data stream modality is less obvious, though: while the class-wise recognition fidelity profits in terms of median intersection over union (IoU), the overall accuracy suffers slightly.

\begin{table}[t!]
\caption{Results on the in-house dataset. The left section presents results for predicting length of stay (LoS) in terms of postoperative hospital days. The right section presents results for predicting complications in terms of CCI. \emph{Pre} and \emph{fine} refer to the use of pretrained and finetuned V-JEPA, respectively, both in the video-only (vid) and multi\-modal (vid+str) setup, as defined in the main text. Shown metrics are: IoU for class~0 (IoU${}_0$; value $\leq$ median), IoU for class~1 (IoU${}_1$; value > median), overall accuracy (Acc).}\label{tab-own}
\centering\setlength{\tabcolsep}{6pt}
\begin{tabular}{@{}lrrr@{}}
\toprule
LoS & \%IoU${}_0$ & \%IoU${}_1$ & \%Acc \\
\midrule
pre (vid) & 46 & 11 & 50 \\
fine (vid) & 46 & \textbf{13} & 50 \\
pre (vid+str) & \textit{57} & 6 & \textit{58} \\
fine (vid+str) & \textbf{61} & \textit{12} & \textbf{63} \\
\bottomrule
\end{tabular}
\quad
\begin{tabular}{@{}lrrr@{}}
\toprule
CCI & \%IoU${}_0$ & \%IoU${}_1$ & \%Acc \\
\midrule
pre (vid) & 55 & \textbf{19} & 59 \\
fine (vid) & 69 & 13 & 70 \\
pre (vid+str) & \textbf{73} & \textbf{19} & \textbf{75} \\
fine (vid+str) & \textit{70} & 15 & \textit{72} \\
\bottomrule
\end{tabular}
\end{table}

\subsubsection{Results on in-house data.} Table~\ref{tab-own} shows results for the tasks of predicting length of stay and complications on the in-house dataset.

Regarding the use of domain-specific finetuning for the underlying V-JEPA video encoder, once more we can see a positive effect, especially when predicting postoperative hospital length of stay. When predicting complications, this effect is less pronounced and even reversed in the multimodal setting. In contrast, the benefit of integrating vital data streams as an additional modality shows a clear benefit here, providing a higher value in almost all metrics.

\section{Discussion and Conclusion}

With our proposed approach for adapting a generic foundation model to domain-specific data and for integrating additional domain-specific data streams in a multimodal setup, we aimed at challenging three~hypotheses.

Regarding our first hypothesis (\emph{a generic foundation model can form the basis for a domain-specific downstream task}), we consider it as shown: even without finetuning V-JEPA, we are on par with the top EndoVis2017 performers in terms of surgical phase recognition accuracy on the HeiCo dataset.

Regarding our second hypothesis (\emph{domain adaptation can increase model performance}), we similarly see a clear trend in the metrics used in our experiments. Future experiments are needed to determine whether the decline in performance on the CCI task after finetuning was a random occurrence or otherwise reveals a systematic underlying root cause.

Regarding our third hypothesis (\emph{the inclusion of additional data sources can increase model performance even further}), the picture is more nuanced: In the HeiCo dataset, where the status of devices in the OR is provided as an additional modality, we can see no clear benefit. This might not be all too surprising though, since the relationship between the provided measurements and the downstream task is a rather indirect one, which makes the measurements unlikely to capture task-relevant information. A similar effect of \emph{modality dominance} has been reported, for example, by Hemker et al.~\cite{hemker_healnet_2024} in datasets where the majority of task-specific information appears to be provided through a single modality. We believe that the same effect is present with the HeiCo data. In our in-house data, the provided additional sensor streams, which contain vital data, appear to be more informative on the outcome of interest, and thus our hypothesis seems to hold there.

In conclusion, our experiments demonstrate that generic foundation models can indeed form a viable basis for solving domain-specific tasks, while the benefit of adding further modalities in a multimodal setup can be rather dependent on their information value for the downstream tasks at hand. By finetuning the V-JEPA video encoder on unlabeled laparoscopic video streams from minimally invasive surgery, we furthermore showed the benefit of domain adaption of such a foundation model. By releasing our model architecture and finetuning weights, we hope to foster future research in surgical data science. 

\begin{credits}
\subsubsection{\ackname}
We would like to thank Julia Ruppel for her assistance in obtaining the necessary patient data. This project was partially funded by the Vontobel Foundation (0867/2024). It was supported by a grant from the Swiss National Supercomputing Centre (CSCS) under project ID lp52 on Alps.

\subsubsection{\discintname}
JAK received a personal research grant from the Department of Surgery, University Hospital Basel. JLL was funded by the Swiss National Science Foundation (P5R5PM\_21766) and the Novartis Foundation for medical-biological Research (\#23C162).
\end{credits}

\bibliographystyle{splncs04}
\bibliography{preprint}

\appendix
\section*{Appendix}\label{sec-app}
\begin{figure}[h!]
\includegraphics[width=\textwidth]{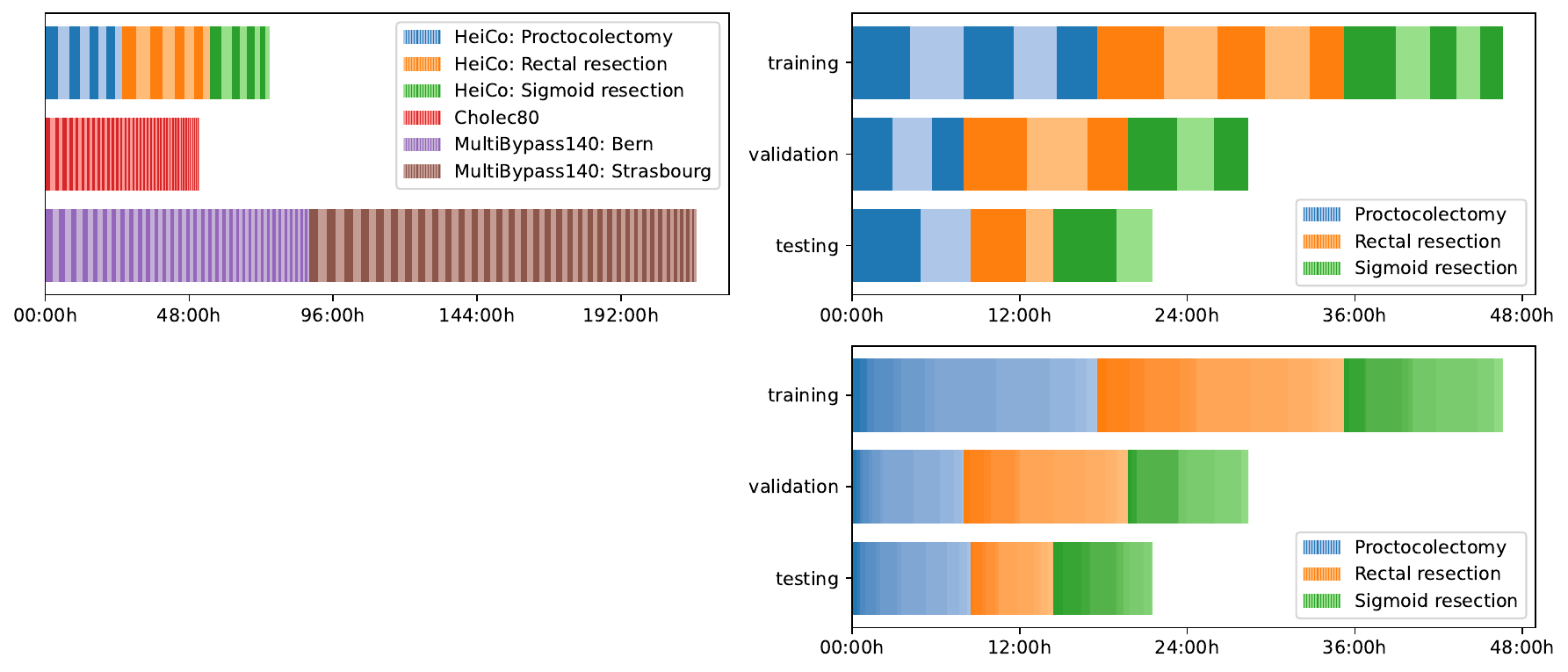}
\caption{Employed public data. Videos by dataset for V-JEPA finetuning \emph{(left)}; videos and splits used for the surgical phase recognition experiment on HeiCo \emph{(right)}.} \label{fig-public}
\end{figure}

\begin{figure}[h!]
\includegraphics[width=\textwidth]{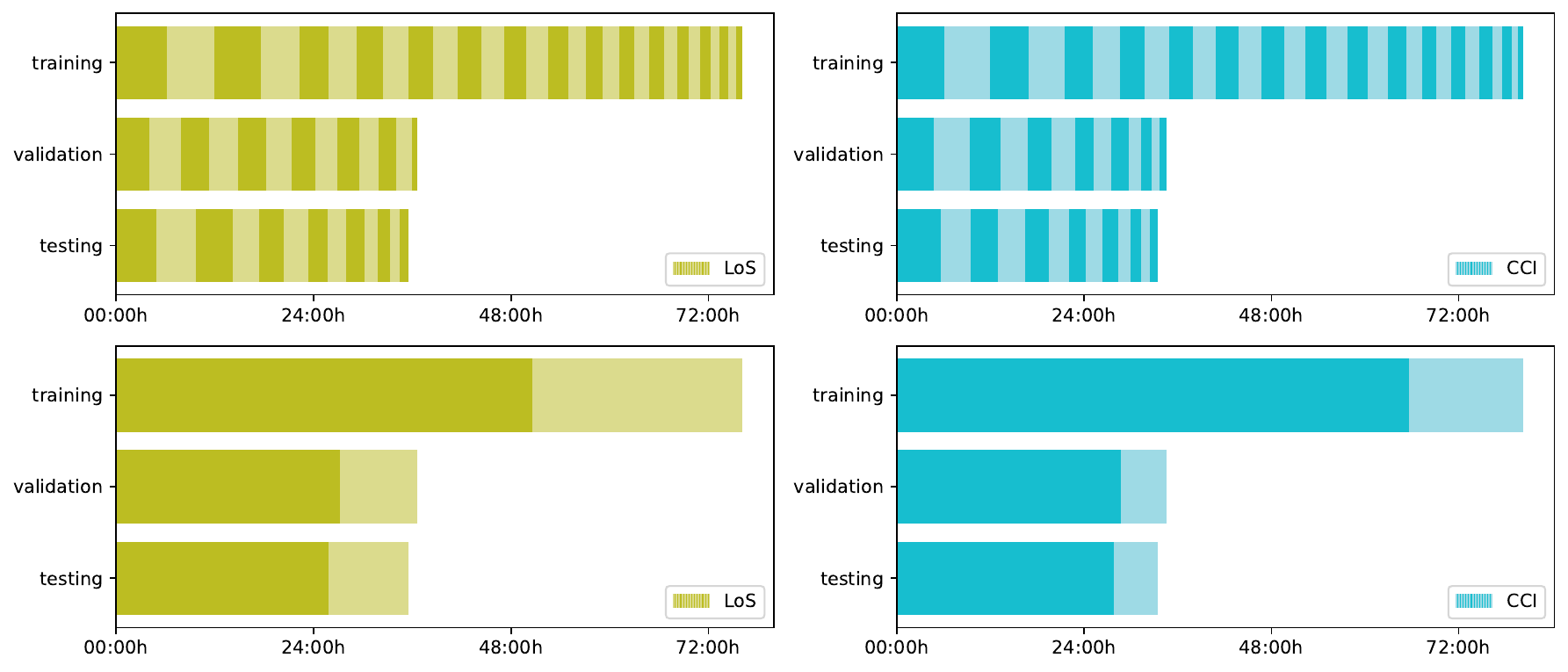}
\caption{Employed in-house data. Videos and splits used for the experiments of predicting length of stay (LoS) in terms of postoperative hospital days \emph{(left)} and  complications in terms of CCI \emph{(right)}.} \label{fig-inhouse}
\end{figure}

The two following figures visualize the extent and composition of the public data (Fig.~\ref{fig-public}) and in-house data (Fig.~\ref{fig-inhouse}) that we used in our experiments. In the top row of each figure, vertical bars in the hatching represent individual videos (sorted by length). In the bottom row of each figure, individual color shades represent individual class labels (sorted in ascending order). No labels were used in V-JEPA finetuning, so no corresponding bar chart is shown there.


\end{document}